\documentclass{bmvc2k}

\pdfoutput=1

\title{Sentence Guided Temporal Modulation for Dynamic Video Thumbnail Generation}

\usepackage{amssymb}
\usepackage{wrapfig}
\usepackage{color, colortbl}

\definecolor{Gray}{gray}{0.9}
\definecolor{LightCyan}{rgb}{0.88,1,1}

\addauthor{Mrigank Rochan}{mrochan@cs.umanitoba.ca}{1}
\addauthor{Mahesh Kumar Krishna Reddy}{kumarkm@cs.umanitoba.ca}{1}
\addauthor{Yang Wang}{ywang@cs.umanitoba.ca}{1,2}

\addinstitution{
 University of Manitoba\\
 Winnipeg, Canada
}
\addinstitution{
 Huawei Technologies\\
 Canada
}

\runninghead{Rochan, Reddy, Wang}{Sentence Guided Video Thumbnail Generation}

\def\etal{\emph{et al}\bmvaOneDot}

\begin{document}

\maketitle

\begin{abstract}
We consider the problem of sentence specified dynamic video thumbnail generation. Given an input video and a user query sentence, the goal is to generate a video thumbnail that not only provides the preview of the video content, but also semantically corresponds to the sentence. In this paper, we propose a sentence guided temporal modulation (SGTM) mechanism that utilizes the sentence embedding to modulate the normalized temporal activations of the video thumbnail generation network. Unlike the existing state-of-the-art method that uses recurrent architectures, we propose a non-recurrent framework that is simple and allows much more parallelization. Extensive experiments and analysis on a large-scale dataset demonstrate the effectiveness of our framework.

\end{abstract}

\section{Introduction}\label{sec:intro}
With the massive rise in videos available online, video thumbnails play a vital role in defining the browsing and searching experience of users. A video thumbnail shows viewers a quick and condensed preview of the entire content in the video \cite{song2016click,liu2015multi,yuan19_acmm}. Viewers often decide whether to watch or skip the video based on its thumbnail \cite{cunningham2008people,song2016click}. Given its importance, there is increasing interest in how to create attractive and expressive thumbnails.

\begin{figure*}[t]
	\center
	\includegraphics[width=1\textwidth]{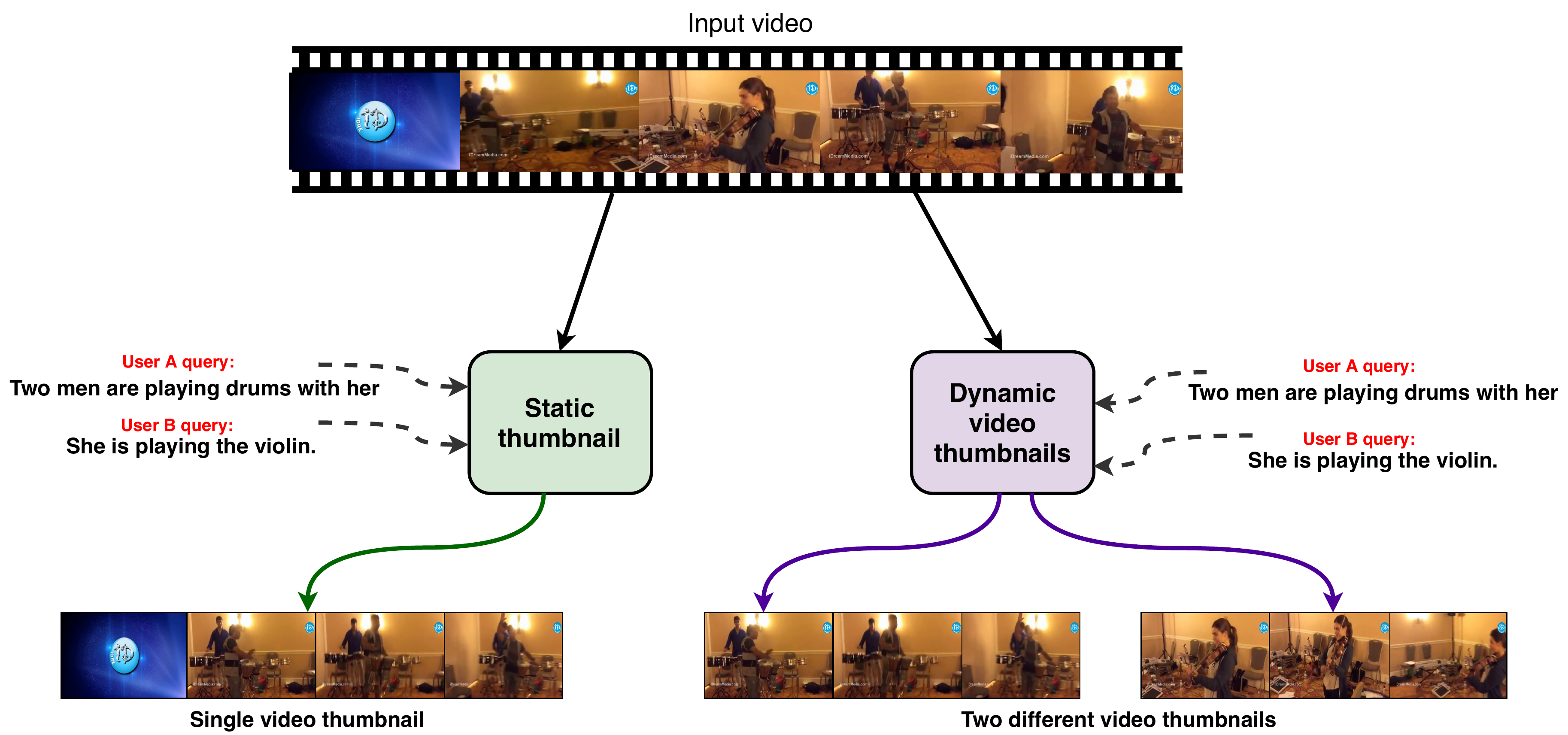}
	\caption{Illustration of the difference between static video thumbnails and sentence specified dynamic video thumbnails. The latter considers the user query sentence when creating the thumbnail.
	}
	\label{fig:static_dynamic}
\end{figure*}

Most of the existing approaches \cite{dirfaux2000key,hasebe2004video,kang2005learn,luo2008towards,song2016click,wang2011extracting} focus on static thumbnail generation, where a thumbnail is generated solely based on the input video. These static thumbnail generation methods overlook rich semantic information such as user query sentence which is usually provided in searching a video. Static thumbnails are not tailored to each viewer's unique interest and may not provide the best online videos browsing experience. Some approaches \cite{liu2011query,liu2015multi} consider the user search query for video thumbnail generation. But they either limit the thumbnail to a single keyframe or confine queries to a single word or a short phrase \cite{yuan19_acmm}. In this paper, we study the recently proposed challenging problem of sentence specified dynamic video thumbnail generation (DVTG) \cite{yuan19_acmm}. Given an input video and a user query expressed as a free form natural language sentence, the goal of DVTG is to generate a video thumbnail that semantically corresponds to the sentence while giving a concise preview of the video. Figure \ref{fig:static_dynamic} shows the difference between static thumbnail generation and DVTG. Despite many potential real world applications (e.g., video search), there is limited work on the DVTG problem.    

The existing state-of-the-art method \cite{yuan19_acmm} for the DVTG task achieves promising results but has some limitations. First, it heavily relies on fine-grained modeling of semantic relationship within the user query sentence and clips or segments of the video \cite{yuan19_acmm}. It does not consider the overall guiding and modulating role that a user query sentence can play in temporally correlating video clips over time. Intuitively, global sentence semantics can serve as a reference in determining and associating sentence-specific video segments over time. Second, the method uses recurrent models (BiGRU \cite{cho2014learning} and pointer networks \cite{vinyals2015pointer}) that are hard to parallelize as they inherently perform sequential computation.

In this paper, we propose a sentence guided temporal modulation (SGTM) mechanism for the DVTG task. We use a self-attention network \cite{wang2018_cvpr} for encoding the user query sentence and adapt a fully convolutional temporal network \cite{ours18_eccv} for the thumbnail generation of an input video. The SGTM mechanism leverages the semantic information from the user query sentence to modulate the normalized temporal activations in the video thumbnail generation network. We also introduce a small auxiliary network to further ensure that the generated video thumbnail semantically corresponds to the given user sentence. Our framework is computationally efficient as its computations are easily parallelizable on GPU architectures. It is also noteworthy that our framework is free from sophisticated multimodal feature fusion (unlike prior method \cite{yuan19_acmm}), making it computationally less complex. We also propose an unsupervised extension of our method that does not need ground-truth video thumbnails during training.

In summary, our contributions in this paper are as follows. (1) We propose a sentence guided temporal modulation (SGTM) mechanism for sentence specified dynamic video thumbnail generation (DVTG). Our method dynamically modulates the temporal activations in video thumbnail generation network using the semantic information of the query sentence. (2) We propose a computationally efficient and simple framework for the DVTG task that offers better parallelization and is free from complex multimodal feature fusion. (3) We propose both supervised and unsupervised version of our method. (4) We conduct extensive experiments and analysis on a large-scale dataset to evaluate our framework. 



\section{Related Work}\label{sec:related}

\noindent{\bf Video thumbnail selection:} Traditional methods \cite{dirfaux2000key,gao2009thematic,hasebe2004video,kang2005learn,mei2009video,song2016click} of video thumbnail generation operate entirely on visual features and characteristics. These methods do not consider any other information related to the video such as textual queries from the user when generating the thumbnail. 
Some recent methods on automatic thumbnail selection \cite{liu2011query,liu2015multi,yuan19_acmm} propose to leverage user textual queries to generate video thumbnails. Most of these methods, however, do not handle complex user queries and are inspired from multi-modal semantic matching model \cite{frome2013devise,pan2014click} that are popular for image search and tagging. The work by Yuan \etal \cite{yuan19_acmm} is the most relevant one to ours. This method combines a variant of pointer network \cite{vinyals2015pointer}, graph convolution network \cite{kipf2017semi} and BiGRU \cite{cho2014learning} to address thumbnail selection based on the user query. This method is computationally complex and prevents parallelization due to sequential computation within. In this paper, we propose an approach that is simple and much more parallelizable.

\noindent{\bf Video highlight and summarization:} Thumbnail generation is related to video highlight detection and video summarization. The goal of video highlight detection is to find the most interesting events or segments in the video \cite{gygli18_acmmm}. Video summarization aims to produce a short and compact overview of the video \cite{gygli18_acmmm}. Although these two tasks are different from thumbnail generation, there is some overlap in techniques employed \cite{song2016click}. Vasudevan \etal \cite{vasudevan2017query} propose a query-adaptive video summarization technique. While promising, this method requires another image annotation dataset to learn the model. Gygli \etal \cite{gygli18_acmmm} use user-generated GIFs to learn video highlight detection. This method can be used to produce thumbnails but they ignore the user query sentence information.

\noindent{\bf Temporal sentence localization in videos:} Temporal sentence localization in video aims to determine the starting and ending of a continuous video segment that matches with the given natural language sentence \cite{anne2017localizing,gao2017tall,liu2018attentive,yuan2019semantic}. Different from this task, video thumbnail generation may contain several nonconsecutive video segments \cite{yuan19_acmm}. Additionally, temporal sentence localization mainly focuses on matching a sentence to a video segment. In contrast, a dynamic video thumbnail should also present a quick preview of the video content along with establishing the relationship with the user query sentence \cite{yuan19_acmm}.

\noindent{\bf Conditional normalization methods:} Our work is also related to conditional batch normalization \cite{de2017modulating} and adaptive instance normalization \cite{huang17_cvpr} methods. These methods spatially normalize the activation from a layer to zero mean and unit variance, and then apply an affine transformation whose parameters are computed using external data \cite{park19_cvpr}. These methods have been successfully applied in image understanding tasks such as visual question answering and image-to-image translation. 

\section{Our Approach}\label{sec:approach}
The input to DVTG consists of a video $V$ and a user query sentence $S$. We denote video by $V=\{v_c\}^C_{c=1}$, where $v_c$ is the feature representation of $c$-th video clip and $C$ is the total number of clips in the video. Similarly, we represent the user query sentence by $S=\{x_n\}^N_{n=1}$, where $x_n$ denotes the word embedding of $n$-th word and $N$ denotes the total number of words in the sentence.  The goal of DVTG is to identify a set of video clips (which may not be consecutive) from $V$ that provide a good preview of the original video $V$ and are semantically consistent with the sentence $S$. 

Given $V$ and $S$, our goal is to learn a mapping function $F(V,S)\in\mathbb{R}^{1\times C\times 2}$, where the output of $F(V,S)$ indicate the scores of whether or not a video clip should be included in the video thumbnail. We call the function $F$ as the sentence guided video thumbnail generation model.


\subsection{Sentence Guided Video Thumbnail Generation Model}\label{sec:model}
\begin{figure*}[h]
\includegraphics[width=1\textwidth]{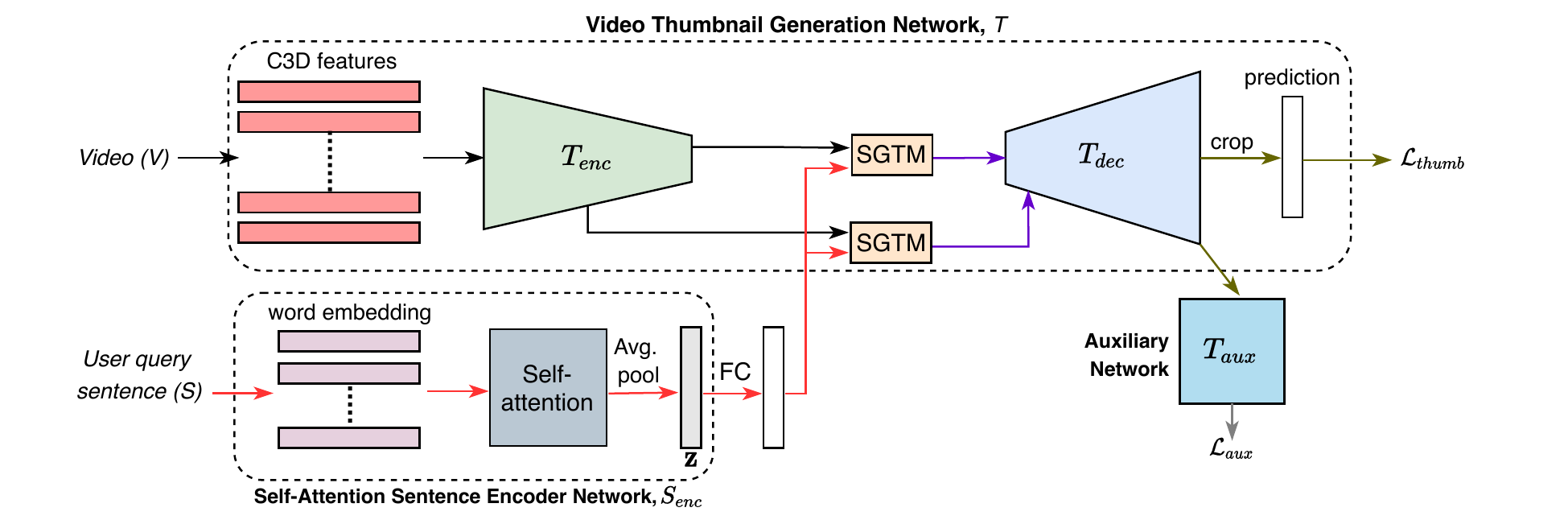}
\caption{An overview of our sentence guided video thumbnail generation model. The model consists of a video thumbnail generation network ($T$), a self-attention sentence encoder network ($S_{enc}$), and an auxiliary network ($T_{aux}$). A sentence guided temporal modulation (SGTM) mechanism is introduced to allow interaction between $T$ and $S_{enc}$. The network $T$ consists of an encoder $T_{enc}$ and a decoder $T_{dec}$. $T$ takes features of video clips in an input video $V$ and predicts whether or not each clip belongs to the video thumbnail. $S_{enc}$ encodes the word-level embedding of user query sentence ($S$) to a vector ${\bf z}$ which is then used by SGTM to modulate the temporal activations from the encoder of $T$ (i.e., $T_{enc}$) to determine sentence-specific video content over time. The role of $T_{aux}$ is to reconstruct ${\bf z}$ to further ensure that the generated video thumbnail aligns well with $S$. We use two losses for learning: a thumbnail generation loss $\mathcal{L}_{thumb}$ on the prediction of $T_{dec}$ and an auxiliary loss $\mathcal{L}_{aux}$ on the output of $T_{aux}$.}
\label{fig:arch}
\end{figure*}

We propose the sentence guided video thumbnail generation model $F$ which consists of three sub-networks, namely a video thumbnail generation network ($T$), a self-attention sentence encoder network ($S_{enc}$) and a auxiliary network ($T_{aux}$). A sentence guided temporal modulation (SGTM) mechanism is proposed to modulate certain activations in $T$ using the output from $S_{enc}$. Figure \ref{fig:arch} shows an overview of our proposed model. In the following, we discuss our model in detail.   
 
\subsubsection{Video Thumbnail Generation Network}\label{sec:vtgn}
The video thumbnail generation network $T$ is an encode-decoder style temporal convolution network. A temporal convolution network mainly performs 1D operations (e.g., 1D convolution, 1D pooling) over time. For instance, given a video with frame-level feature representations, this network can operate over frames enabling it to capture the temporal dependencies among the frames. In this paper, we adapt FCSN \cite{ours18_eccv} which is a form of encoder-decoder based temporal convolution network designed for video summarization.

The input to $T$ is a sequence of clip-level feature representations of video $V$, i.e., $V \in \mathbb{R}^{1 \times C \times D_c}$ where $C$ is the total number of clips and $D_c$ is the dimension of the feature representation of each video clip. The output of $T$ is of dimension $1 \times C \times 2$ which denotes the scores of each clip being a part of the thumbnail or not.

$T$ consists of an encoder $T_{enc}$ and a decoder $T_{dec}$. $T_{enc}$ has seven convolutional blocks. Each of the first five convolution blocks ($conv1$ to $conv5$)  consists of multiple temporal convolution and ReLU operations, with a max pooling at the end. The last two blocks ($fc6$ and $fc7$) has a temporal convolution, a ReLU and a dropout operation. $T_{enc}$ produces two feature maps as outputs, one from the last layer and another one as a skip connection from $conv4$ block. The outputs of $T_{enc}$ is fed to $T_{dec}$ as inputs. The first input to  $T_{dec}$ goes through a $1\times1$ convolution and a temporally fractionally-strided convolution ($deconv1$) which is then combined with the second input after applying a  $1\times1$ convolution to it. Lastly, it has another temporally fractionally-strided convolution ($deconv2$) and a crop operation so as to obtain the final prediction of dimension  $1 \times C \times 2$.


\subsubsection{Self-Attention Sentence Encoder Network}\label{sec:sen}
The self-attention sentence encoder network ($S_{enc}$) is responsible for encoding the user query sentence $S$ to a fixed length vector $\bf{z}$. 

Self-attention \cite{parikh16_self,vaswani2017_nips} has been shown to be a powerful technique in natural language processing. We apply the non-local model \cite{wang2018_cvpr,zhang2019_icml} with self-attention mechanism in $S_{enc}$. This enables $S_{enc}$ to effectively model relationships between different words in the sentence. 

We represent each word in the sentence using its word embedding vector. The sentence $S$ can be written as $S \in \mathbb{R}^{D_w \times N}$, where $N$ is the number of words in the sentence and $D_w$ is the dimension of word embedding. The attention between two words can be computed as:

\begin{equation}
\Phi_{j,i} = \frac{\exp(A_{ij})}{\sum_{i=1}^{N}\exp(A_{ij})}, \; \; \text{where} \;\; A_{ij}=f_1(x_i)^Tf_2(x_j),
\label{eqn:attn}
\end{equation} 
Here $\mathit{f_1}$ and $\mathit{f_2}$ represent two distinct feature spaces, i.e., $\mathit{f_1}(x_i) = W_{f_1}x_i$ and $\mathit{f_2}(x_j) = W_{f_2}x_j$. The attention score $\Phi_{j,i}$ denotes the extent with which $i$-th word is related to $j$-th word. The output of self-attention is $\Lambda = (\Lambda_1, \Lambda_2, ..., \Lambda_N) \in \mathbb{R}^{D_w \times N}$, where
\begin{equation}
\Lambda_{j} = \sum_{i=1}^{N} \Phi_{j,i}h(x_i), \; \; \text{where} \;\; h(x_i) = W_{h}x_i .
\label{eqn:self-attn}
\end{equation}
Note that $W_{f_1} \in \mathbb{R}^{d \times D_w}$, $W_{f_2} \in \mathbb{R}^{d \times D_w}$ and $W_{h} \in \mathbb{R}^{D_w \times D_w}$ are the learnable weights implemented using $1 \times 1$ convolutions. In our experiments, we set $d = D_w/8$ . Lastly, we apply average pooling on the resultant self-attended features $\Lambda$ to obtain the vector ${\bf z} \in \mathbb{R}^{D_w}$ representing the whole sentence $S$.


\subsubsection{Sentence Guided Temporal Modulation}\label{sec:sgtm}
In order to generate a video thumbnail that corresponds to a user sentence, it is necessary to establish the relationships between the clips of video and the sentence. The rich semantics in the sentence provides a strong signal to temporally associate it with the video clips. Motivated by this, we propose the sentence guided temporal modulation (SGTM) mechanism. In practice, SGTM is similar to temporal-adaptive instance normalization \cite{rochan2020adaptive} that extends adaptive instance normalization \cite{huang17_cvpr} (initially designed for images) to videos. SGTM plays a key role in our proposed model. It allows the interaction between the video thumbnail generation network $T$ (Sec. \ref{sec:vtgn}) and the self-attention sentence encoder network $S_{enc}$ (Sec. \ref{sec:sen}) so as to generate a video thumbnail that closely relates to the given user query sentence. SGTM uses the semantic information in the sentence to guide and temporally modulate the features of the input video fed to the video thumbnail generation network $T$.

Let ${\bf z}$ be the sentence vector representation from $S_{enc}$ and $A \in \mathbb{R}^{1 \times M \times C}$ be the activations from one specific layer in the temporal convolution network $T$, where $M$ is the temporal length of the activation and $C$ is the number of channels. We firstly forward ${\bf z}$ to a fully-connected layer (FC) to generate a vector of length $2C$. We use this generated vector to obtain two modulation vectors $\alpha \in \mathbb{R}^{C}$ and $\beta \in \mathbb{R}^{C}$. Next, we use these modulation vectors to modulate the each channel of activation map $A$. We can express the modulated activation map at $m \in M$ and $c \in C$ as $\hat{A}_{m, c}$ which is computed as:
\begin{equation}
\hat{A}_{m, c} = \alpha_c \cdot \frac{A_{m,c} - \mu_c(A)}{\sigma_c(A)} + \beta_c,
\label{eq:sgtm}
\end{equation}
where $\mu_c$ and $\sigma_c$ are the channel-wise mean and standard deviation calculated independently for each input video along the temporal length of the activation map $A$. 

From Eq. \ref{eq:sgtm}, we can infer that a normalized temporal activation map with zero mean and unit variance in each channel is subjected to an affine transformation (scale and shift) whose values are predicted using the sentence representation ${\bf z}$. Note that the affine transformation is temporally invariant and there are no learnable parameters involved in this mechanism. By using this translation-based design, our goal is to allow the sentence semantics information to modulate each temporal feature map of the input video.

In our model, we apply SGTM to the two output activations of the encoder $T_{enc}$ (see Fig. \ref{fig:arch}). We use the sentence representation ${\bf z}$ from $S_{enc}$ to produce two sets of affine parameter vectors ($\alpha_i, \beta_i$ where $i = 1, 2$) using a FC layer. The parameters $\alpha_i$ and $\beta_i$ correspond to the $i$-th SGTM. 

\subsubsection{Auxiliary Network}\label{sec:auxnet}
To further ensure that the generated thumbnail aligns well with the query sentence, we introduce an auxiliary network ($T_{aux}$) next to the decoder $T_{dec}$ of $T$ (Sec. \ref{sec:vtgn}). It is a small network whose input is the output of $deconv2$ layer in $T_{dec}$ which is forwarded to a $1 \times 1$ convolution and an average pooling operation so as to reconstruct the user sentence vector representation ${\bf z}$ learned by $S_{enc}$ (Sec. \ref{sec:sen}).

\subsection{Learning and Optimization}\label{sec:learning}
Our learning objective includes a thumbnail loss and an auxiliary loss.

\noindent{\bf Thumbnail loss}: For an input video $V$ with $C$ clips and the ground-truth binary indicator label vector denoting whether a clip belongs to thumbnail or not, we define a cross-entropy loss $\mathcal{L}_{thumb}$ on the prediction of the video thumbnail generation network $T$ as:

\begin{equation}\label{eq:cls_loss}
\mathcal{L}_{thumb} = -\frac{1}{C}\sum_{c=1}^{C} \log\Bigg(\frac{\exp(\delta_{c,l_{c}})}{\sum_{j=1}^{2}\exp(\delta_{c,j})}\Bigg) ,
\end{equation}
where $\delta_{c,j}$ is the predicted score of $c$-th video clip to be labeled as $j$-th class (non-thumbnail or thumbnail) and $\delta_{c,l_c}$ is the score for the ground-truth class $l_c$ for the $c$-th video clip. 

\noindent{\bf Auxiliary loss}: This loss aims to minimize the difference between the sentence representation ${\bf z}$ from $S_{enc}$ and the reconstructed sentence representation $\hat{\bf z}$ from $T_{aux}$. We define the reconstruction loss $\mathcal{L}_{aux}$ as:

\begin{equation}\label{eq:aux_loss}
\mathcal{L}_{aux} = ||{\bf z} - \hat{\bf z} ||^2,
\end{equation}
where ${\bf z},\hat{\bf z} \in \mathbb{R}^{D_w}$ and $||\cdot||$ denotes the $L_2$ norm.

\noindent{\bf Final loss}: We define the final loss $\mathcal{L}_{final}$ as:

\begin{equation}\label{eq:final_loss}
\mathcal{L}_{final} = \mathcal{L}_{thumb} + \mathcal{L}_{aux}.
\end{equation} 

The aim of the learning is to find the optimal parameters $\Theta_{T}^*$, $\Theta_{S_{enc}}^*$ and $\Theta_{T_{aux}}^*$ of the networks $T$, $S_{enc}$ and $T_{aux}$, respectively. We can express the learning objective as follows:
\begin{equation}\label{eq:goal}
\Theta_{T}^*,\Theta_{S_{enc}}^*,\Theta_{T_{aux}}^* =  \underset{\Theta_{T},\Theta_{S_{enc}},\Theta_{T_{aux}}}{\arg\min} \; \mathcal{L}_{final}(T, S_{enc}, T_{aux}) .
\end{equation}

For simplicity, we denote our proposed sentence guided dynamic video thumbnail generation model (Sec. \ref{sec:model}) learned from Eq. \ref{eq:goal} by \texttt{Guided-DVTG}.

In summary, \texttt{Guided-DVTG} captures the global information of the video through its video thumbnail generation network (Sec. \ref{sec:vtgn}) which is crucial to generate thumbnails that provide overall content preview. It also has the ability to dynamically modulate its prediction using the user sentence via the sentence guided temporal modulation (Sec. \ref{sec:sgtm}). Moreover, it further ensures semantic correspondence with the sentence using the auxiliary network (Sec. \ref{sec:auxnet}). As a result, \texttt{Guided-DVTG} can produce dynamic video thumbnails that provide a quick preview of the video while satisfying the user query sentence.

\section{Experiments}\label{sec:exp}

\subsection{Setup}
\noindent\textbf{Dataset:} We conduct experiments on the sentence specified dynamic video thumbnail generation dataset by Yuan \etal \cite{yuan19_acmm}. This dataset is based on the ActivityNet Captions dataset \cite{krishna2017dense}. It has $10, 204$ video-sentence pairs where each pair is labeled with $4$ video thumbnail annotations. The thumbnail annotation for each video is at clip-level where each clip is of $2$ seconds and with no more than $5$ clips from the video included in the final video thumbnail. $70\%$ of the dataset is used for training, $15\%$ for validation and the remaining $15\%$ for testing.

\noindent\textbf{Feature representation:} We follow prior work \cite{yuan19_acmm} for video and word embedding representation. We evenly split every video in the dataset into $2$ second clips and represent each clip with the C3D features \cite{tran15_iccv} provided by the ActivityNet Challenge 2016. For each word, we obtain a $300$ dimensional word embedding using Glove \cite{socher14_glove}.  

\noindent\textbf{Training details:} We train all our models from scratch with a constant learning rate $0.0001$ using the Adam optimizer \cite{kingma15_iclr}. During training, for a video-sentence pair, we find the most consistent video thumbnail annotation among the $4$ thumbnail annotations and treat it as the ground-truth annotation. However, in testing, we evaluate the predicted video thumbnail by comparing it against all the $4$ thumbnail annotations. Note that similar process is followed by previous work \cite{yuan19_acmm}.   

\noindent\textbf{Evaluation metrics:} Following prior work~\cite{yuan19_acmm}, we use the F1 and IoU scores to measure the performance of our methods. These metrics measure the agreement between the generated thumbnail and the ground-truth thumbnail annotations. A higher value is desirable on these metrics. 

\subsection{Main Results and Comparisons}
In addition to comparing with prior state-of-the-art video thumbnail generation methods, we also define several strong baselines that are as follows:

\noindent\textbf{FCSN} \cite{ours18_eccv}: This is a state-of-the-art model in video summarization task. We extend it to create our video thumbnail generation network $T$ (see Sec. \ref{sec:vtgn}). We directly train and evaluate FCSN on the dataset in this paper. Note that this is a generic video thumbnail model that does not consider the user query sentence.


\noindent\textbf{IN-FCSN}: This baseline is a variant of video thumbnail generation network $T$ where we replace the proposed SGTM mechanism (see Sec. \ref{sec:sgtm}) with the temporal instance normalization layer \cite{ulyanov2016instance} with learnable affine transformation parameters. Note that we do not have the networks $S_{enc}$ and $T_{aux}$ in this model. This results in another generic thumbnail model.

\noindent\textbf{IN-FCSN-concat}: We obtain this baseline when we train video thumbnail generation network $T$ by concatenating the user sentence vector representation (obtained by averaging the words embedding) with the video-clip features of the input video. We again replace SGTM mechanism with temporal instance normalization layer with learnable affine transformation parameters. This results in a dynamic video thumbnail generation model as it incorporates user sentence information in the model. 

\begin{figure*}[h]
	\center
	\includegraphics[width=1\textwidth]{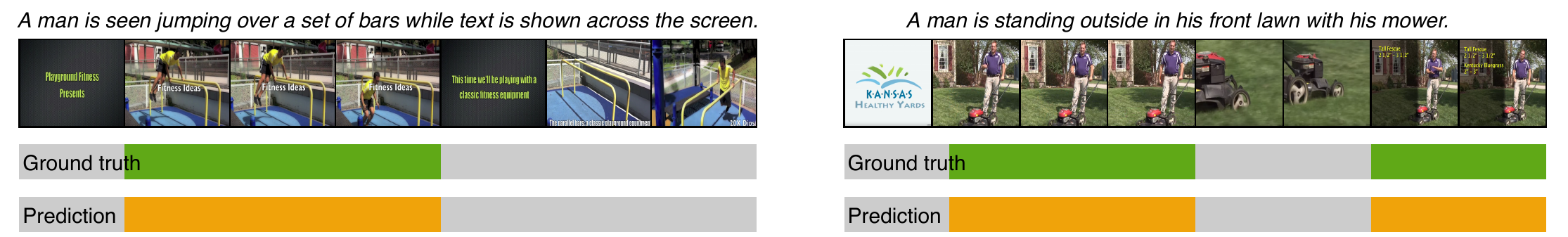}
	\caption{Example qualitative results produced by \texttt{Guided-DVTG}. The gray, green and orange bars indicate the video length, ground-truth and thumbnail predictions, respectively.
	}
	\label{fig:qual}
\end{figure*}

In Table \ref{table:main_results}, we compare our final model \texttt{Guided-DVTG} with the prior and baseline methods. We outperform the baselines and other alternative methods except GTP \cite{yuan19_acmm}. Unlike GTP, we do not use sophisticated multimodal feature fusion. Moreover, GTP uses recurrent models that perform sequential computation within training samples which prevents parallelization, whereas our model is completely non-recurrent that allows much more parallelization. Figure \ref{fig:qual} shows example video thumbnails generated by our \texttt{Guided-DVTG} model.
\begin{table*} [h]
\centering
\begin{tabular}{c|c|c}
	\hline
	Method & F1 & IoU \\
	\hline
	Random & 0.3604 & 0.2379\\
	RankNet \cite{gygli16_cvpr} & 0.4013 & 0.2770 \\
	VSEM \cite{liu2015multi} & 0.4386 & 0.3098 \\
	QARE \cite{vasudevan2017query} & 0.4285  & 0.2986 \\
	CTRL \cite{gao2017tall} & 0.4303 & 0.3084 \\
	ACRN \cite{liu2018attentive} & 0.4456  & 0.3271 \\
	\rowcolor{Gray}
	GTP \cite{yuan19_acmm} & 0.5285  &  0.3933 \\
	\hline
	FCSN \cite{ours18_eccv} (ours) & 0.4295 & 0.3101  \\
	IN-FCSN (ours) & 0.4426 &  0.3140 \\
	IN-FCSN-concat (ours) & 0.4286 &  0.3084 \\
	\hline
	\rowcolor{LightCyan}
	\texttt{Guided-DVTG} (ours) & 0.4758 & 0.3405\\
	\hline
\end{tabular}	
\caption{Performance comparison (in terms of F1 and IoU) between \texttt{Guided-DVTG} and other alternative methods. Results of previous methods are taken from \cite{yuan19_acmm}. Best and second best methods are highlighted in gray and cyan, respectively.
}
\label{table:main_results}
\end{table*}

\subsection{Analysis}

\noindent{\bf Role of Modulation Parameters:} We analyze the importance of modulation parameters $\alpha_c$ and $\beta_c$ (Eq. \ref{eq:sgtm}) in the SGTM mechanism (Sec. \ref{sec:sgtm}). Table \ref{table:analyse_mod_params} compares the performance for different possible solutions of these parameters. We find that when these parameters are predicted ($\alpha_c$=$\alpha_c^s$, $\beta_c$=$\beta_c^s$) from another network using the user query sentence (i.e., in \texttt{Guided-DVTG}), the model achieves much better performance as compared to cases when they are set to fixed values ($\alpha_c$=1, $\beta_c$=0) or directly learned ($\alpha_c$=$\alpha_c^*$, $\beta_c$=$\beta_c^*$) in the main video thumbnail generation network (Sec. \ref{sec:vtgn}). Therefore, the proposed SGTM is key to dynamic video thumbnail generation.
\begin{table}[h]
\small
\centering
	\begin{tabular}{c|c|c|c}
	\hline
	Method & $\alpha_c=1$, $\beta_c$=0  & $\alpha_c$=$\alpha_c^*$, $\beta_c$=$\beta_c^*$ & $\alpha_c$=$\alpha_c^s$, $\beta_c$=$\beta_c^s$\\
	\hline
	IN-FCSN  & 0.4062 (0.2904) & 0.4426 (0.3140) & - \\
	IN-FCSN-concat  & 0.4480 (0.3192)  & 0.4286 (0.3084) & - \\
	\texttt{Guided-DVTG} & -  & - & \textbf{0.4758} ({\bf 0.3405}) \\
	\hline
	\end{tabular}
	\caption{Impact of modulation parameters on video thumbnail generation. Here we indicate F1 and IoU (in bracket) for different solutions of parameters $\alpha_c$ and $\beta_c$ in Eq. \ref{eq:sgtm}.}
	\label{table:analyse_mod_params}
\end{table}

\noindent{\bf Impact of Auxiliary Network:} We study the impact of auxiliary network (Sec. \ref{sec:auxnet}) and the loss $\mathcal{L}_{aux}$ (Sec. \ref{sec:learning}) associated with it in learning the thumbnail model. In order to verify their contribution, we remove them from our final model \texttt{Guided-DVTG} and perform learning. We call the learned model \texttt{Guided-DVTG-NA} and compare the performance in Table \ref{table:ablation}(a). We notice a drop in performance which highlights the importance of the auxiliary network and its loss in our \texttt{Guided-DVTG} model.

\noindent{\bf Unsupervised Guided-DVTG:} We develop an unsupervised variant of our \texttt{Guided-DVTG} model. When we remove the supervised loss $\mathcal{L}_{thumb}$ (Sec. \ref{sec:learning} and Eq. \ref{eq:final_loss}) and perform learning only using the auxiliary loss $\mathcal{L}_{aux}$ which is completely unsupervised, we obtain the unsupervised version of our model to which we call \texttt{Guided-DVTG$_{unsup}$}. In Table \ref{table:ablation}(b), we compare its performance with the state-of-the-art unsupervised method, BeautThumb \cite{song2016click}, when evaluated on the dataset in this paper. Our model \texttt{Guided-DVTG$_{unsup}$} significantly outperforms BeautThumb \cite{song2016click}. This result is very appealing since gathering labeled video thumbnail data is extremely expensive. 

\begin{table*}[h]
	\small
	\centering
	\begin{tabular}{ cc }   
	\begin{tabular}{ c|c|c } 
	\hline
	Method & F1  & IoU \\
	\hline
	\texttt{Guided-DVTG}  & \textbf{0.4758} & \textbf{0.3405}\\
	\texttt{Guided-DVTG-NA}  & 0.4644 & 0.3296 \\
	\hline
	\end{tabular} &  
	\begin{tabular}{ c|c|c } 
	\hline
	Method & F1  & IoU \\
	\hline
	Random & 0.3604 & 0.2379\\
	BeautThumb \cite{song2016click}  & 0.3837 & 0.2544\\
	\texttt{Guided-DVTG$_{unsup}$}  & \textbf{0.4222} & \textbf{0.2835} \\
	\hline
	\end{tabular}\\
	(a) & (b) \\ 
	\end{tabular}
\caption{(a) Impact of auxiliary network and its loss. (b) Performance comparison of unsupervised methods. Result of BeautThumb \cite{song2016click} is taken from \cite{yuan19_acmm}.}
\label{table:ablation}
\end{table*}


\section{Conclusion}\label{sec:conclude}
In this paper, we have proposed a simple yet effective framework for the DVTG task. At the core of our framework is the proposed SGTM mechanism that modulates the normalized temporal activations in the video thumbnail generation network to effectively correlate sentence-specific video clips over time. Instead of applying recurrent neural architectures, we propose a non-recurrent solution that offers much more parallelization on GPU hardware. Our proposed framework does not involve complex multimodal feature fusion commonly used in vision-language tasks such as DVTG. The experimental results and analysis on a large-scale dataset demonstrate that our proposed method achieves superior or competitive performance against the state-of-the-art methods.

\section{Acknowledgements}
The authors acknowledge financial support from NSERC and UMGF funding. We thank NVIDIA for donating some of the GPUs used in this work.

\bibliography{rochan}
\end{document}